\pdfoutput=1

\documentclass[11pt]{article}

\usepackage[final]{acl}

\usepackage{times}
\usepackage{latexsym}
\usepackage{xcolor}
\usepackage{float}
\usepackage{subcaption}
\usepackage{listings}
\usepackage{changepage}
\usepackage{tcolorbox}
\usepackage{amsmath}
\usepackage[T1]{fontenc}

\usepackage{graphicx}
\usepackage{xcolor}

\usepackage[utf8]{inputenc}

\usepackage{microtype}

\usepackage{inconsolata}

\usepackage{graphicx}

\title{Quantifying reliance on external information over parametric knowledge during Retrieval Augmented Generation (RAG) using mechanistic analysis}

\usepackage{hyperref}
\author{
  \textbf{Reshmi Ghosh\textsuperscript{1}},
  \textbf{Rahul Seetharaman\textsuperscript{2}},
  \textbf{Hitesh Wadhwa\textsuperscript{2}},
  \textbf{ Somyaa Aggarwal\textsuperscript{2}},
\\
  \textbf{Samyadeep Basu\textsuperscript{1,3}},
  \textbf{Soundararajan Srinivasan\textsuperscript{1}},
  \textbf{Wenlong Zhao\textsuperscript{2}},
 \textbf{Shreyas Chaudhar\textsuperscript{2}},
\\
  \textbf{Ehsan Aghazadeh\textsuperscript{2}}
\\
  \textsuperscript{1}Microsoft,
  \textsuperscript{2}University of Massachusetts, Amherst,
 \textsuperscript{3}University of Maryland, College Park,
}

\begin{document}
\maketitle
\begin{abstract}

Retrieval Augmented Generation (RAG) is a widely used approach for leveraging external context in several natural language applications such as question answering and information retrieval. Yet, the exact nature in which a Language Model (LM) leverages this non-parametric memory or retrieved context isn't clearly understood. This paper \textbf{\textit{mechanistically}} examines the RAG pipeline to highlight that LMs demonstrate a  ``shortcut'' effect and have a strong bias towards utilizing the retrieved context to answer questions, while relying minimally on model priors. We propose (a) Causal Mediation Analysis; for proving that parametric memory is minimally utilized when answering a question and (b) Attention Contributions and Knockouts for showing the last token residual stream do not get enriched from the subject token in the question, but gets enriched from tokens of RAG-context. We find this pronounced ``shortcut'' behaviour to be true across both LLMs (e.g.,LlaMa) and SLMs (e.g., Phi).

\end{abstract}

\section{Introduction}
Retrieval Augmented Generation (RAG) \cite{lewis2021retrievalaugmented}is a popular method to enhance a Language Model's (LLM) capability to reason and execute tasks by leveraging additional context provided during inference time \cite{shao2023eragent}\cite{singh2023domain}\cite{ingestai2023rag}. Additionally, researchers have also explored shortcomings of RAG systems, such as inconsistent responses\cite{liu-etal:2023:arxiv} and only~\cite{wu2024faithful} delved into the balance between a model’s internal knowledge and externally retrieved information, examining their practical value. 

Several research papers have proposed the approaches for editing knowledge in language model, including techniques like ROME \cite{meng2022locating}, MEMIT \cite{meng2022mass} to update or correct facts. On the flip side, with the popularity of LLM integration for various tasks leveraging properitary, enterprise, and private data, the use of RAG framework has increased to tackle \textit{hallucinations} while reasoning on \textbf{new never seen before} (out of distribution) knowledge tasks. However, a comprehensive study mechanistically probing of Langauge Model's behavior of choosing between information from RAG-generated context over intrinsic parametric knowledge has not been conducted to the best of our knowledge.

\section{Probing Methods}

To mechanistically interpret the knowledge contributions towards factual reasoning by LLMs and SLMs, we use three methods for causal mediation, described as follows:
\textbf{Causal Tracing} \cite{meng2022locating} identifies specific hidden states that significantly influence factual predictions. The approach involves a clean run, corrupted run and a corrupted-with-restoration run. The corrupted run involves corrupting a certain span of the text, and running the forward pass of the model. In the restoration run, activations from the clean run are patched one by one into the corrupted run, and the increase in answer probability is observed; the most crucial activations are thus causally determined. The causal importance of a certain activation is quantified using Indirect Effect, which is defined as the difference between the corrupted run and the corrupted-with-restoration run probabilities: $
\text{IE}(h^{(l)}_i) = P^*_{\text{clean}}(h^{(l)}_i)[y] - P^*[y]$. The Average Indirect Effect of a hidden node is an average of IE over all the prompts in the dataset. 

The \textbf{Attention Contribution} \cite{yuksekgonul2024attention}, focuses on the role of attention mechanisms in shaping the output of language models. This approach  investigates how attention weights, particularly from the subject token in a query to the last token position, contribute to the model's predictions. By examining the norm of these attention weights \( \|a^{(\ell)}_{i,T}\| \), we observe what tokens the last token pays the most attention to, during the generation process. The \textbf{Attention Knockout} mechanism \cite{geva2023dissecting} identifies critical attention edges in transformer-based models that are essential for maintaining prediction quality. The process involves identifying critical edges whose removal significantly degrades the model's prediction quality, by means of setting the attention from position $i$ to $j$ in the attention matrix to $-\infty$. 
\section{Data and Models}
For examining model activations for causal tracing, patching and inspecting AIE, whilesystematically analyzing attention contributions we choose open source LMs like LlaMa-2 (7B) and Phi-2 (2.7B) models. And for understanding the behavior in the non-RAG setting, we leverage the \textit{Knowns 1000 dataset}, a dataset of 1209 prompts \cite{meng2022locating}. For the RAG setting, we augment the \textit{Knowns 1000 dataset} with added context generated synthetically using GPT-4. We use GPT-4 generated context to control the length of each segment within the RAg-context and also the presence of \textit{\textbf{attribute}} or \textit{\textbf{object}}.
\section{Results}
Experimenting with LLaMa and Phi-2 family of models on 1209 samples from the knowns fact dataset for vanilla-case and RAG-scenario, demonstrate that both models exhibit a strong bias towards utilizing external knowledge provided by RAG.

Utilizing \textit{Causal Tracing} method and measuring Average Indirect Effect (AIE) at different positions of the prompt, such as Last Subject Token (LST), Last Token (LT) , it is found that for the vanilla-case (non-RAG) LST had high AIE, but it substantially lowered when RAG-generated context was added. As concluded in \cite{meng2022locating}, LST has the largest influence from model priors and lowering AIE of LST demonstrates reduced influence of parametric memory. We specifically observe $\sim$10X decrease in AIE of LST for LLaMA-2 and $\sim$35X decrease in AIE of LST for Phi-2, when RAG-generated context is added.

This finding was further corroborated by utilizing two other probing methods - Attention Knockouts and Attention Contributions. The LT is a crucial component in the LLM decoding process, as it is projected onto the vocabulary during decoding time. Thus any information that has to be decoded, will be propagated by the MLP and attention layers to the LT residual stream. We measured the attention contributions from the Subject Token (ST) to the LT and observe a substantial decrease in ST contributions for the RAG-scenario as compared to the vanilla non-RAG case where no external context is provided. For LlaMa-2, the mean attention contribution decreased by $\sim$1.6X for RAG case, in comparison to non-RAG vanilla case, and for Phi-2 a reduction of 7x was observed for ST contribution. Conversely, the \textit{answer token} contribution for RAG setting, increases significantly for LlaMa-2 and Phi-2 in comparison to ST contribution in the RAG setting. This further confirms our hypothesis of the LLM being less reliant on its parametric memory and exhibiting a "shortcut" behavior.

Using Attention knockouts \cite{geva2023dissecting} approach, it is observed that "knocking out" attention from ST to the LT, reduces the probability of the answer in the LM's last token predictions by 20\%  in LlaMa-2 and 25\% in Phi-2. This is in sharp contrast to the RAG setting, where knocking off attention at ST positions leads to <5\% drop in the answer probabilities. This finding further reinforces the finding that the model takes a "shortcut" while relying minimally on its parametric memory.

\section{Conclusions and Future Work}

Using Causal Tracing \cite{meng2022locating} in over 1200 samples of the \textit{known facts dataset}, in RAG-scenario for LLaMa-2 and Phi-2, we observe a reduced AIE on the last subject token, and potentially reduced dependence on parametric memory. This is further corroborated by our experiments with attention contributions and attention knockouts. Using three mechanistic probing techniques, we observe 1) reduced reliance on parametric memory 2) reduced information flow from the subject token to the last token residual stream 3) a shortcut behavior where information from the attribute token flows to the last token residual stream during factual predictions in the RAG setting.

Future work will address the extension to larger LMs (> 13B parameters). We also plan to study the impact of LM behavior in longer context, and in settings where language models are known to exhibit primacy and recency bias \cite{liu-etal:2023:arxiv} in a future work. Additionally, we aim to replicate our findings using a conventional RAG pipeline to automatically create context rather than synthetically generating it using GPT4.

\bibliography{acl_latex}

\end{document}